# On Stacked Denoising Autoencoder based Pre-training of ANN for Isolated Handwritten Bengali Numerals Dataset Recognition

**Al Mehdi Saadat Chowdhury[1,*], M. Shahidur Rahman[1], Asia Khanom[2], Tamanna Islam Chowdhury[2], Afaz Uddin[2]**

[1] Department of Computer Science and Engineering, Shahjalal University of Science & Technology, Sylhet, Bangladesh.

[2] Department of Computer Science and Engineering, North East University Bangladesh, Sylhet, Bangladesh.

Email: saadat_cse@yahoo.com, rahmanms.bd@gmail.com, akhanom8@gmail.com, tamannachy092@gmail.com, afazuddinahmed22@gmail.com

**Abstract**: This work attempts to find the most optimal parameter setting of a deep artificial neural network (ANN) for Bengali digit dataset by pre-training it using stacked denoising autoencoder (SDA). Although SDA based recognition is hugely popular in image, speech and language processing related tasks among the researchers, it was never tried in Bengali dataset recognition. For this work, a dataset of 70000 handwritten samples were used from (Chowdhury and Rahman, 2016) and was recognized using several settings of network architecture. Among all these settings, the most optimal setting being found to be five or more deeper hidden layers with sigmoid activation and one output layer with softmax activation. We proposed the optimal number of neurons that can be used in the hidden layer is 1500 or more. The minimum validation error found from this work is 2.34% which is the lowest error rate on handwritten Bengali dataset proposed till date.

## 1. INTRODUCTION

Artificial intelligence tasks like vision, language, speech and robotics has recently gained much interest among researchers due to their inherent complexity. One such task is the recognition of isolated handwritten numerals in which Artificial Neural Network (ANN) based recognition showed better performance, especially when the network is shallow having only a single or two layer of hidden neurons. In reality though, a shallow ANN is not efficient in terms of the number of computation units, thus in terms of required examples (Bengio and LeCun, 2007). Researchers believed that the composition of several levels of nonlinearity would be key to efficiently model complex relationships (Vincent et. al., 2010). Thus the deep neural network research becomes a major research area from times.

Contrary to the belief, deep networks continuously showed poor performances due to their problematic non-convex optimization. Training a deep network to optimize the supervised criterion directly using gradient descent with random initialization does not seem to work very well. This scenario was changed dramatically by the seminal work of Hinton et al. (2006) and Hinton and Salakhutdinov (2006), where they suggested to train the network in a local unsupervised fashion first (which is called pre-training) and then fine tune the network using the supervised data. For pre-training Restricted Boltzmann Machine was used in their work. This network was called Deep Belief Network. Following their work, several other similar methods was proposed over time (Lee et. al., 2008; Ranzato et. al., 2007). One such method is the stacked denoising autoencoder based pre-training method which was proposed by Vincent et al. (2010).

Several works including Basu et. al. (2005), Bashar et. al. (2004) and Khan et. al. (2004) were done previously to recognize Bengali handwritten numerals using artificial neural network, but all these works didn't show good performances due to the use of very small dataset and complex feature set based recognition. Listed accuracy in these articles were also inaccurate, since no steps were taken to remove the overfitting problem.

---

* Corresponding author: saadat_cse@yahoo.com

In this work, we have used stacked denoising autoencoder (SDA) which was proposed by Vincent et al. in (2010) to recognize isolated handwritten Bengali numerals. For recognition, a dataset similar to MNIST containing 70,000 handwritten Bengali numeral samples was first collected and prepared as described in the work of the same author (Chowdhury and Rahman, 2016). As found in that article, traditional MLP based recognition can't past a single or two layer margin. Hence, this SDA based work not only solves the limitation of an MLP, but also it provides better classification accuracy on the test set. This work is the first instance of such SDA based deep learning framework ever tried on handwritten Bengali numerals. As with other machine learning algorithms, the major goal is not to devise a new algorithm every time for every single work, rather to find the optimal parameter settings for a particular algorithm applied to a task, this work also made an exhaustive grid search in the parameter space to find the most optimal setting for applying deep SDA based recognition on Bengali handwritten digits.

The rest of the paper is organized as follows. Section 2 explains the ins and outs of a stacked denoising autoencoder. Section 3 discuss about the results of the network training and the justification of the choices made in the work. This section also includes the performance accuracy found from various operation applied to the network. The article concludes with some final remarks listed in section 4.

## 2. STACKED DENOISING AUTOENCODER

In this section, an introduction to an autoencoder is given first, followed by a denoising autoencoder and finally how several denoising autoencoder can be stacked together to form the final network is explored.

### *2.1 Autoencoder*

A traditional autoencoder is a three layer network that has an input layer *(x)*, a fully connected encoding hidden layer *(y)* and another fully connected decoding layer *(z)*. Fig. 1 demonstrates the network architecture of an autoencoder.

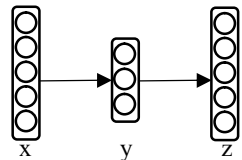

Fig. 1: A traditional Autoencoder

The encoding layer encodes the input $x \in [0,1]^d$ (which is a vector of d-dimension) through a deterministic mapping $y = f_\theta(x) = s(\omega^T x + b)$, where the parameter vector is $\theta = \{\omega, b\}$. This encoded input (of a different dimension than the original input vector) $y \in [0,1]^{d'}$ is further reconstructed in the decoding layer through another mapping $z = g_{\theta'}(y) = s(\omega'^T y + b')$, where the parameter vector is now $\theta' = \{\omega', b'\}$. The dimension of the reconstructed layer must be equal to the dimension of the input layer, hence $z \in [0,1]^d$. Sometimes the weight matrix is constrained to $\omega' = \omega^T$ which in this instance is called *tied weights*. The choice of the deterministic function *s(.)* is a decision choice to make during training.

### *2.2 Denoising Autoencoder*

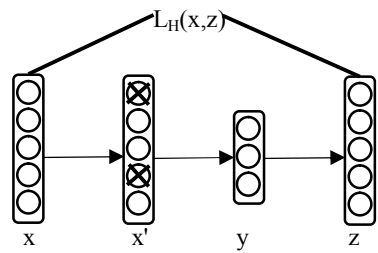

Fig. 2: A denoising autoencoder

A denoising autoencoder (DA) is just an extended version of the traditional autoencoder as shown in fig. 2. In DA, the input is first corrupted by adding some random noise to it. This corruption can be done by making some of the input node to zero, letting the reconstruction layer guess what the input was. In this way, the autoencoder cannot learn identity weights, which in turn provides better learning. $L_H(x,z)$ is the cross-entropy loss between the input $x$ and the reconstructed output $z$ that is used for updating weights in the gradient descent algorithm.

### 2.3 Stacked Denoising Autoencoder

Similar to the stacking of Restricted Boltzmann Machine as proposed in Hinton et al. (2006) and Hinton and Salakhutdinov (2006), stacking denoising autoencoder also initializes a deep network with appropriate weights. In this case, input corruption only takes place at the beginning; later layer is trained on uncorrupted version of the input. A logistic regression layer is put on top of the encoding layers, thus a supervised fine tuning criterion can be achieved using this architecture by applying stochastic gradient descent optimization algorithm.

## 3. PROPOSED PARAMETER SETTINGS AND RESULTS

In this section, we listed the test results found from our training of a neural network with pre-training by an SDA. The main target here is to find the optimal open parameters for the recognition task.

### 3.1 Activation for the Output Layer

For the output layer, we need an activation that would be bounded. Literature suggests that the bounded activations are sigmoid, tanh and softmax. Sigmoid's output is in the range of [0,1], tanh's output is in the range [-1,1] and softmax's output is in the range [0,1]. The major difference between sigmoid and softmax is, all of the softmax's output for a single training example sums to 1, thus softmax acts like a probability distribution. Thus softmax becomes a natural choice for output activation of our network. Previous work from the same author (Chowdhury and Rahman, 2016) also justifies the choice.

### 3.2 Activation for the Hidden Layer

Unlike choosing softmax for the output layer, we cannot choose the activation for hidden layer simply, because the hidden layer is computing the internal representation, rather than computing the final output. Hence, an exhaustive search through the parameter space is required for finding the optimal activation in the hidden layer. Fig. 1 listed the outcome of the experiment. The experiment clearly indicates that, the sigmoid activation is the most appropriate activation for hidden layer.

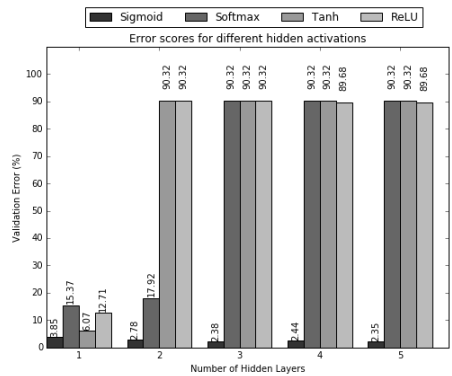

Fig. 3: Optimal Activation Functions for Hidden Layer

### 3.3 Optimal Number of Hidden Layers and Neurons

Along with providing information about the optimal hidden activation, Fig. 3 also gives data on the optimal number of hidden layers. As we can see, increasing layers also increases performance accuracy of the network. For this work, we have tested up to five layers. The reason for not going beyond is the amount of training time needed to complete running a single algorithm is more than 8 hours even with an NVIDIA

GeForce 750Ti graphics card and Core i5 Processor, 4GB DDR4 RAM setting. These algorithms were run under Linux environment, data and the execution of code happens inside the graphics card instead of the CPU and RAM, and for parallelism, CUDA was used.

To find the optimal number of hidden neurons, the network is trained further by setting the output activation to softmax and hidden activation to sigmoid and the result is listed in Table 1.

Table 1: Validation error on different hidden neurons

| No. of Hidden Layers | Hidden Neurons | Validation Error (%) |
|---|---|---|
| 4 | 300 | 2.88 |
|  | 500 | 2.62 |
|  | 700 | 2.40 |
|  | 1000 | 2.44 |
|  | 1500 | 2.37 |
| 5 | 300 | 2.96 |
|  | 500 | 3.70 |
|  | 700 | 2.50 |
|  | 1000 | 2.35 |
|  | 1500 | **2.34** |

Although the behavior of the network seems a bit stochastic, one exciting trend is though that, increasing number of neurons minimizes the error bar more. Thus, at the fifth layer with 1500 hidden neurons, we are getting the minimum classification error which is **2.34%**.

### *3.4 Comparison with Other Models*

For evaluation, we have compared our SDA based model with some well-established models whose results are listed in Table 2. All these models are implemented using Pythons scikit-learn module, running on their generalized default settings.

Table 2: Finding optimal number of hidden neurons

| Models | Validation Error (%) |
|---|---|
| Logistic Regression | 19.41 |
| Decision Tree | 27.42 |
| K-Nearest Neighbors (K=3) | 12.73 |
| Nearest Centroids | 30.53 |
| Gaussian Naïve Bayes | 35.18 |
| Multinomial Naïve Bayes | 27.3 |
| Bernoulli Naïve Bayes | 30.65 |
| Support Vector Machine (SVM) | 16.65 |
| SVM with Linear Kernel | 16.99 |
| SVM with RBF Kernel | 16.65 |
| SVM with Sigmoid Kernel | 89.82 |
| ANN with a single hidden layer (hidden activation: ReLu, output activation: softmax, hidden neuron: 700, No pre-training envoloved) | 3.95 |
| Proposed Method | **2.34** |

It is apparent from Table 2 that, all other models fall short compared to our proposed model on Bengali handwritten digit classification.

### 4. CONCLUSION

In this work, we have used an isolated Bengali handwritten digit dataset proposed by (Chowdhury and Rahman, 2016) for recognizing them using artificial neural network based architecture pre-trained by a stacked denoising autoencoder. The main target of this research is to find the optimal parameter settings of this well-known architecture. We have tested from one to five hidden layers with sigmoid, softmax, tanh and rectified linear unit activation and found that the softmax activation for output unit, sigmoid activation for hidden unit, five hidden layers with 1500 neurons is the best setting for the recognition of Bengali handwritten digit dataset. We also conclude that, the network train past five layers with more hidden units may perform even better, but such a large system is expensive in terms of time to implement. A minimum validation error of 2.34% has been found in the process.